\documentclass[twocolumn,10pt]{asme2ej}

\usepackage{graphicx} 
\usepackage{hyperref}   
\hypersetup{
	colorlinks=true,
	linkcolor=blue,
	citecolor=blue,
	urlcolor=blue,
}
\usepackage[square,numbers]{natbib}
\usepackage{url}
\usepackage{amsmath}
\usepackage{amssymb}
\usepackage{booktabs}
\usepackage{multirow}
\usepackage{overpic}
\usepackage[table]{xcolor} 

\usepackage{fancyhdr}

\title{Representation Learning for Sequential Volumetric Design Tasks}

\renewcommand{\headrulewidth}{0pt} 

\author{Md Ferdous Alam
    \affiliation{
        Department of Mechanical Engineering\\
	The Ohio State University\\
	Columbus, OH 43210 \\
    Email: alam.92@osu.edu
    }	
}
\author{Yi Wang 
    \affiliation{Principal Research Scientist \\ Autodesk Research\\
        Email: yi.wang@autodesk.com
    }
    
}

\author{Chin-Yi Cheng 
    \affiliation{Google Research\\
        Email: hajimecheng@gmail.com
    }
    
}

\author{Jieliang Luo 
    \affiliation{Sr. Principal AI Research Scientist\\ Autodesk Research\\
    Email: rodgerljl@msn.com
    }
    why do why 
}

\begin{document}

\maketitle    

\begin{abstract}
Volumetric design, also called massing design, is the first and critical step in professional building design which is sequential in nature. As the volumetric design process requires careful design decisions and iterative adjustments, the underlying sequential design process encodes valuable information for designers. Many efforts have been made to automatically generate reasonable volumetric designs, but the quality of the generated design solutions varies, and evaluating a design solution requires either a prohibitively comprehensive set of metrics or expensive human expertise. While previous approaches focused on learning only the final design instead of sequential design tasks, we propose to encode the design knowledge from a collection of expert or high-performing design sequences and extract useful representations using transformer-based models. Later we propose to utilize the learned representations for crucial downstream applications such as design preference evaluation and procedural design generation. We develop the preference model by estimating the density of the learned representations whereas we train an autoregressive transformer model for sequential design generation. We demonstrate our ideas by leveraging a novel dataset of thousands of sequential volumetric designs. Our preference model can compare two arbitrarily given design sequences and is almost $90\%$ accurate in evaluation against random design sequences. Our autoregressive model is also capable of autocompleting a volumetric design sequence from a partial design sequence.
\end{abstract}


\section{Introduction}
\label{sec:intro}

\begin{figure}
    \centering
    \includegraphics[width=0.5\textwidth]{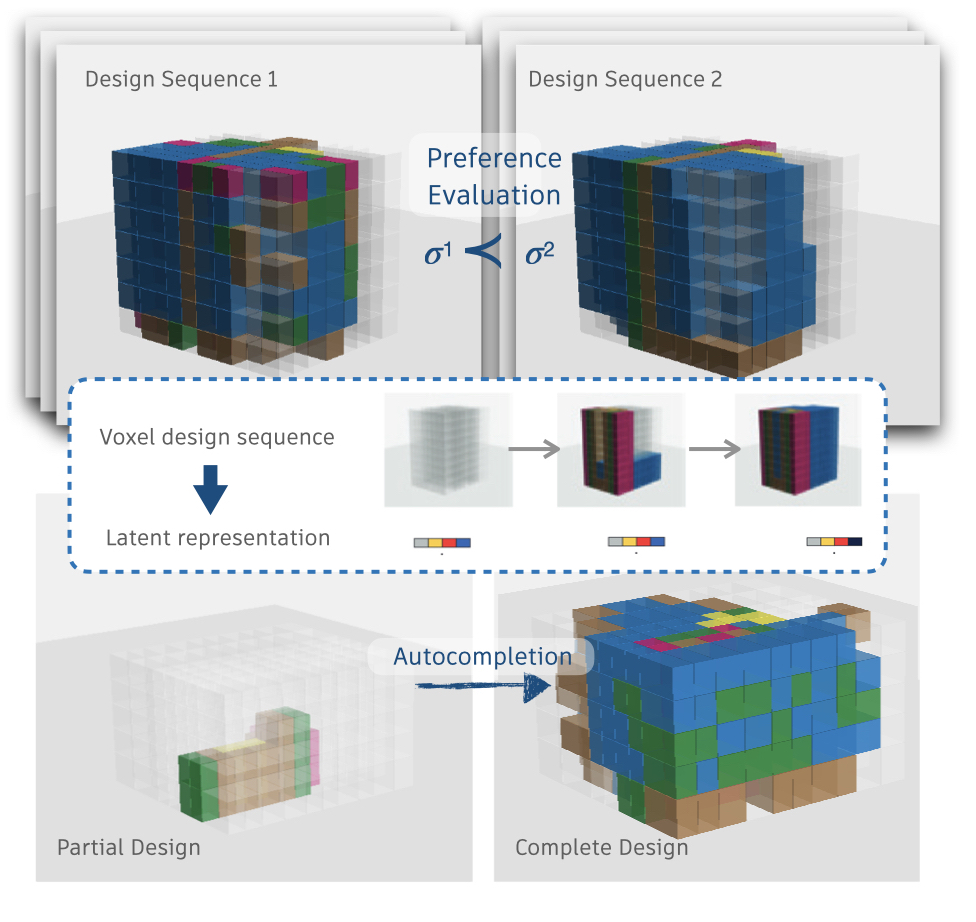}
    \caption{In this paper, we conduct representation learning for high-dimensional sequential volumetric design, and apply the learned representation for two downstream tasks: 1) preference evaluation over two design sequences (above) and 2) auto-completion with a partial design (below). }
    \label{fig:teaser_image}
    \vspace{-0.25cm}
\end{figure}

Many design tasks are essentially sequential in nature, which requires numerous iterations and is often time-consuming due to the use of heuristics or manual approaches \cite{Peters, rahman2021predicting, shergadwala2018quantifying}. We argue that the sequential approach to engineering design is particularly useful and convenient not only from the technical perspective but also from the design perspective. The sequential approach can allow human designers to understand, intervene and modify an existing design and thus have the potential to accelerate the design pipeline by a large margin. In this study, our core motivation is to develop neural network models that can learn complex sequential design representations and thus accelerate downstream design automation tasks. While heuristic methods are the industry standard in engineering design, our learning-based sequential design systems can provide two major benefits over heuristics and manual approaches; 1) sequential generation of novel designs that can be used by a human expert designer for early design motivations or prototyping, and 2) creating design automation tools such as design evaluation. As a motivating example of sequential design, we demonstrate our ideas on architectural design which requires iterative modifications to match desired specifications. Additionally, there is a lack of tools that can help a designer with some initial realistic designs to make design iterations fast \cite{nauata2020house, chang2021building}. Here we explore a learning-based approach for evaluating and generating complex design tasks. We specifically focus on the challenging task of sequential volumetric designs, a crucial step in professional building design. Volumetric design is a complex process that requires a multitude of manual efforts from expert designers. Several efforts have been made to automatically generate reasonable volumetric designs~\cite{chang2021building}. Unfortunately, the quality of the generated design solution varies, and evaluating a design solution requires either a prohibitively comprehensive set of metrics or expensive human expertise. Here we aim to develop neural network models that can learn the underlying latent representation for such design tasks from a collection of high-performing sequential volumetric design data. We argue that this perspective on data-driven modeling of the sequential design procedure has several advantages due to the following reasons. First, from the workflow perspective, the existing solutions usually don't reveal the decision-making process, which creates a barrier to human-AI interaction. Second, this idea opens up new possibilities to build AI-assisted sequential generative design tools that can incorporate expert feedback to fine-tune the design. Inspired by the success of Transformer-based models in the field of natural language processing (NLP)~\cite{brown2020language, devlin2018bert} and computer vision (CV)~\cite{bao2021beit, DosovitskiyB0WZ21}, we make the first endeavor to explore the idea of latent representation learning for sequential volumetric design, where the inputs are a sequence of voxel-based representations. Our key motivation is to use the learned representations for several crucial downstream tasks, such as reconstruction, preference evaluation, and auto-completion. We utilize self-attention layers from transformer models for these sequential tasks. Our contributions can be summarized as the following:

\begin{enumerate}
    \item We present a novel `sequential volumetric design' dataset and develop a self-attention-based encoder-decoder model to learn the latent representation of this dataset
    \item We demonstrate the effectiveness of the learned representations for autocompletion and reconstruction of design sequences
    \item We introduce and develop a preference model architecture that can provide preference over two volumetric design sequences by combining representation learning and flow-based density estimation
    \item Finally, our approach shows how the sequential approach can provide unique benefits to engineering design process
\end{enumerate}


\section{Related Work}
\label{sec:related-work}

Volumetric design and our proposed framework for learning representations span several topics such as 3D representation learning, density estimation, and high-dimensional sequential representation learning.  

 \paragraph{3D Representation Learning} Recent literature shows extensive use of learning-based approaches for 3D design and representation learning \cite{shu20203d, ranade2022activationnet, li2022predictive, li2023design}. 3D shapes are commonly rasterized and processed into voxel grids for analysis and learning \cite{liu2021survey, williams2019design, cunningham2020sparsity}. Due to the correspondence and similarity between voxels and 2D pixels, many works have explored voxel-based classification and segmentation using volumetric convolution \cite{yan2018second, ye2020hvnet, zhou2018voxelnet}. However, the volumetric convolution suffers from capturing rich context information with limited receptive fields. Recently, Transformer-based 3D backbones have proved to be a more effective architecture as long-range relationships between voxels can be encoded by the self-attention mechanism in the Transformer modules \cite{mao2021voxel, he2022voxel}. Apart from 3D object detection and recognition, voxel-based latent generative models also have shown success for 3D shape generation \cite{sanghi2022clip, wu2016learning, zhou20213d}. However, voxel grids are memory-intensive. With the increase in dimensionality, voxel grids grow cubically, and thus, are expensive to scale to high resolution. Recently, interest in CAD-based representation learning has emerged due to the accessibility of large-scale datasets including collections of B-reps and program structure \cite{lambourne2021brepnet, jones2022self}, CAD sketches\cite{Li:2020:Sketch2CAD}, and CAD assemblies\cite{willis2020fusion, willis2021joinable}. 

\paragraph{Density Estimation}

Density estimation is heavily utilized in generative models to learn the probability density of a random variable, $p(x)$, given independent and identically distributed (i.i.d.) samples from it. Although density estimation allows sampling from a distribution or estimating the likelihood of a data point, this becomes challenging for high-dimensional data. Some popular approaches for high-dimensional density estimation include variational auto-encoder (VAE) \cite{kingma2013auto}, autoregressive density estimator \cite{germain2015made}, and flow-based models \cite{dinh2016density}. VAEs do not offer exact density evaluation and are often difficult to train. Autoregressive density estimators are sequential in nature and are trained autoregressively by maximum likelihood. An alternative approach, that we utilize in this study, is the normalizing flows which model the density as an invertible transformation of a simple reference density. 

\paragraph{High-dimensional Sequential Representation Learning} 
Representation learning for engineering design is studied somewhat extensively in the literature \cite{valdez2022latent, xu2015machine, wang2022three}. On the other hand,  sequential representation learning tasks such as video frames, are challenging and have been widely studied with the transformer-based mechanism. With the prominent success of CLIP (Contrastive Language-Image Pre-training)~\cite{pmlr-v139-radford21a}, several methods have adopted and extended CLIP to video representation ~\cite{luo2022clip4clip, portillo2021straightforward, bain2021frozen}. Learning meaningful state representations from visual signals is also essential for reinforcement learning (RL). Several prior works have explored the use of auxiliary supervision in RL to learn such representations ~\cite{guo2020bootstrap, jaderberg2016reinforcement}. Recently, ~\cite{yu2022mask, zhu2022masked} leverage transformer-based models to learn representation from consecutive video frames by using mask-based latent representation as the auxiliary objective, which can further improve the sample efficiency of RL algorithms.

\paragraph{Building Layout Generation \& Analysis} 
Although neural networks have been widely used in architectural design previously \cite{rhee2023three}, prior work on learning-based 3D building layout generation is rare. To the best of our knowledge, Building-GAN~\cite{chang2021building} is the only study exploring this direction where they train a graph-conditioned generative adversarial network (GAN) with GNN and pointer-based cross-model module to produce voxel-based volumetric designs. On the 2D layout generation, House-GAN~\cite{nauata2020house} also proposes a graph-conditioned GAN, where the generator and discriminator are built on relational architecture. Similarly, ~\cite{di2020end} uses multiple GAN modules to generate interior designs with doors, windows, and furniture. Layout-GMN~\cite{patil2021layoutgmn} uses an attention-based graph-matching network to predict structural similarity. Transformer-based neural networks have also been studied in this field. HEAT~\cite{chen2022heat} can reconstruct the underlying geometric structure with a given 2D raster image. HouseDiffusion~\cite{shabani2022housediffusion} uses a diffusion model for vector-floorplan generation.

\section{Methodology}
\label{sec:methodology}

\subsection{Data Synthesis}

\begin{figure}
    \centering
    \includegraphics[width=0.50\textwidth]{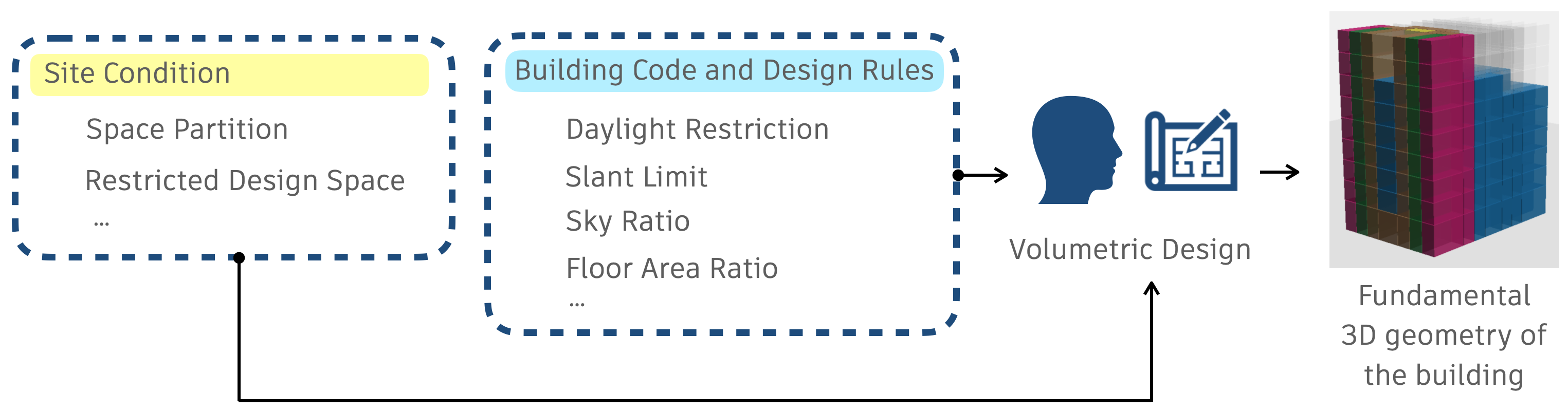}
    \caption{As the first critical step in professional building design, volumetric design takes into consideration of site condition, building code, and design rules (left). The outcome is a 3D structure where different colors represent different room types (right).}
    \label{fig:volumetric-design}
\end{figure}

The volumetric design process usually performs in a given valid design space with several constraints, such as FAR and TPR\footnote{Floor area ratio (FAR) is derived by dividing the total area of the building by the total area of the parcel. Target program ratio (TPR) defines the approximate ratio between different room types.}. The partition of the design space depends on the architect's decisions or statistical heuristics. However, large collections of architectural volumetric designs are prohibitively expensive. For this reason, we opted for a heuristic agent built on top of \cite{chang2021building} that can procedurally create a volumetric design. First, we generate different site conditions randomly based on rules and knowledge provided by professional architects. The heuristic agent creates a design based on the site conditions and then computes the associated program graph and constraints. Later, an action sequence is extracted from these program graphs that can sequentially create a valid design sequence. Finally, this action sequence is fed into Building-Gym, an OpenAI gym-like interface~\cite{brockman2016openai}, that can produce a sequence of design states if actions are provided. Building-Gym sequentially provides the next state of the volumetric design based on the action taken at each timestep. Each design state, $s_t$, is a $10 \times 10 \times 10$ non-uniformed voxel space where each voxel has two channels to represent a) size and b) room type. We define seven room types: elevators, stairs, mechanical rooms, restrooms, corridors, office rooms, and non-existent rooms. For visualization purposes, each room type is color-coded. For example, as shown in Fig. \ref{fig:volumetric-design}, blue denotes office rooms and red denotes elevators. The color-coding details can be found in the Appendix. Our main motivation for developing Building-Gym is to provide a powerful sequential voxel building design simulator for researchers. Our initial dataset consists of $10^4$ optimal action sequences where each action sequence interacts with Building-Gym and produces a sequence of valid voxel design spaces that follow the design principles. An action sequence is of arbitrary length and can be represented by $(a_0, a_1, ..., a_{T})$ where each action $a_t$ represents the location, size, and room type for each new voxel within the voxel space. In this way, using each action Building-Gym can provide us with a new design state. Next, we utilize these optimal action sequences to interact with Building-Gym and create $10^4$ sequences of voxel design spaces. Note that each sequence is a collection of high-performing voxel design spaces where each final state produces a diverse and valid voxel design space of a building. Fig. \ref{fig:data-generation} shows how Building-Gym gradually creates a valid voxel design of a building using a sequence of actions. 

\subsection{Voxel design space to vector representation}
\begin{figure}
    \centering
    \includegraphics[width=0.475\textwidth]{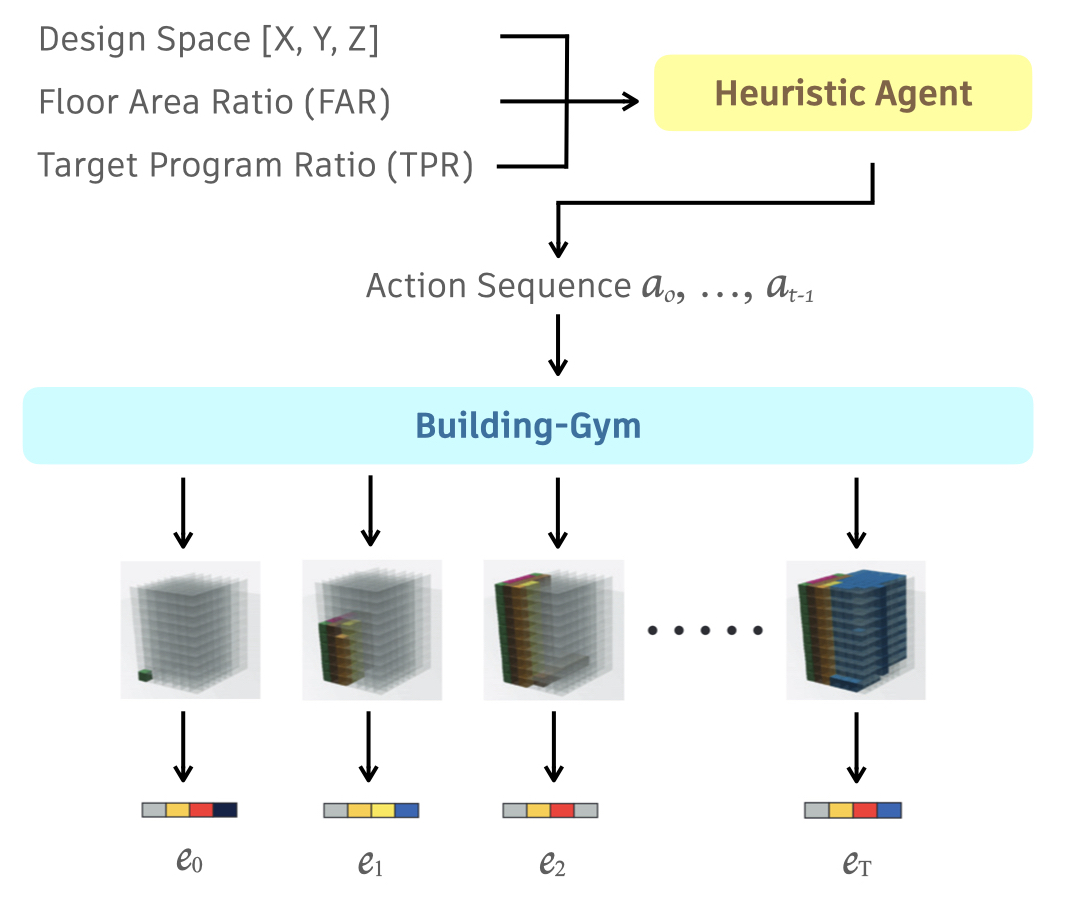}
    \caption{The pipeline to generate design embeddings: 1) our heuristic agent takes in a group of design constraints and outputs an action sequence for each design; 2) our Building-Gym environment takes in the action sequence and outputs a sequence of voxel-based design states, which are flattened into design embeddings.}
    \label{fig:data-generation}
\end{figure}

To convert the voxel design spaces into neural network-friendly representations each state in the sequences is flattened into a $1000$-dimensional vector. We coin this design-to-vector representation as ``design embedding'', $\mathbf{e}_t \in \mathbb{R}^{1000}$, which can be directly fed into the neural networks. Design embedding can be thought of as an analogy to word-to-vector representation in natural language processing. The process of generating one sequence of design embeddings is shown in Fig. \ref{fig:data-generation}. In the initial dataset $\mathcal{D}$, each design sequence has a length between $100$ and $810$. The distribution of the sequence length within $\mathcal{D}$ is shown below in Fig. \ref{fig:data_stat}. We found out that designs that are too short are not meaningful enough for any model to learn from. Thus we remove very short-length sequences from the dataset which results in a total number of $7296$ sequences. Based on our experiments, this dataset contains a diverse range of building designs and is sufficient to learn the sequential correlations in volumetric building design. We consider $6612$ sequences for training all the models reported in this paper and the other $684$ sequences for evaluation of the models. Finally, we sample design states from each sequence at a fixed frequency to truncate the sequence length into a more manageable value e.g. maximum length of $82$ instead of $800$. This sampling makes each sequence more interpretable by keeping the sequence size small and also makes the data processing pipeline more manageable. 

\begin{figure}[!htb]
    \centering
    \includegraphics[width=0.50\textwidth]{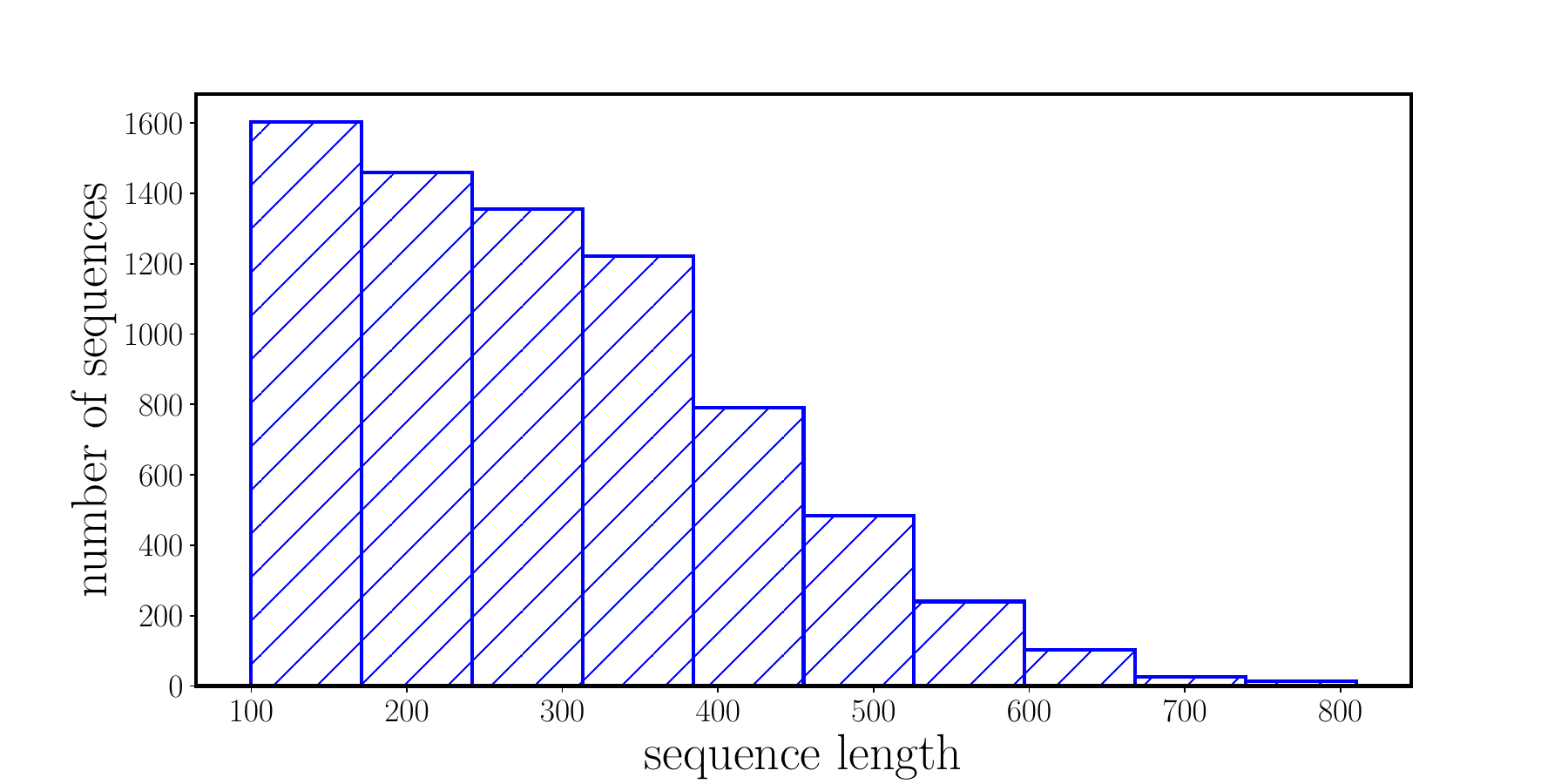}
    \caption{Distribution of sequence lengths in our dataset}
    \label{fig:data_stat}
\end{figure}

\subsection{Representation Learning for Sequential Design}

Our main idea for learning effective representations of sequential design data is to estimate the density of the underlying latent space so that we can both sample data from the distribution and obtain the probability of a single data point. One straightforward way to estimate the density is to use likelihood-based models such as variational autoencoders \cite{kingma2013auto}. In the case of VAE, the parameters of the posterior are jointly learned with the parameters of the decoder by minimizing the evidence lower bound (ELBO) loss: 
\begin{equation}\label{eq:elbo}
\small \mathcal{L}(\theta, \phi) = \mathbb{E}_{z\sim q_\theta(\mathbf{z}|\mathbf{e})}\left[ \log(p_\phi(\mathbf{e}|\mathbf{z})) \right] - \text{KL}(q_\theta(\mathbf{z}|\mathbf{e})||p(\mathbf{z}))
\normalsize
\end{equation}

\noindent where, $\mathbf{z}$ is the latent representation of the design embedding, $\mathbf{e}$, $p_\phi(\cdot)$ is the decoder parameterized by $\phi$, $q_\theta(\cdot)$ is the approximate posterior distribution parameterized by $\theta$, and $p(\mathbf{z})$ is the prior latent distribution. We consider a traditional zero mean and unit variance Gaussian prior, meaning $\mathbf{z} \sim \mathcal{N}(\mathbf{0}, \mathbf{I})$ for each of the latent vectors of the sequence. One challenge with the likelihood-based model is that the regularization term on the right side of Eq. \ref{eq:elbo}, often leads to poor accuracy of the model. To overcome such challenges, we propose to decouple density estimation from learning the latent representation. Thus we propose a two-step procedure; 1) learn the latent representation of the input sequences, and 2) estimate the density of the learned latent space. 

Due to the sequential nature of the data, step 1 can be achieved using multi-head self-attention models such as transformers \cite{vaswani2017attention, radford2019language}. To this end, we again consider two types of approaches for training the transformer-based models. The first approach is motivated by computer vision tasks where we try to reconstruct each design embedding in the sequence. Note that our goal is to reconstruct a whole sequence, unlike the traditional approaches where only a single design is reconstructed for learning the latent representations. In this way, the transformer model would take an input sequence of design embeddings $(\mathbf{e}_0, \mathbf{e}_1, \dots, \mathbf{e}_T)$ and predict $(\hat{\mathbf{e}}_0, \hat{\mathbf{e}}_1, \dots, \hat{\mathbf{e}}_T)$ where $\mathbf{e}_t$ is the $t$-th design embedding of the sequence. Our second approach is motivated by NLP tasks, where we consider autoregressively predicting each design embedding sequentially. The objective can be expressed as maximizing the conditional probability, $\text{arg max}_\theta p(\mathbf{e}_t|\mathbf{e}_0, \mathbf{e}_1, \dots, \mathbf{e}_{t-1}; \theta)$, with respect to some parameter $\theta$.

\begin{figure*}[!htb]
    \centering
    \includegraphics[width=0.875\textwidth]{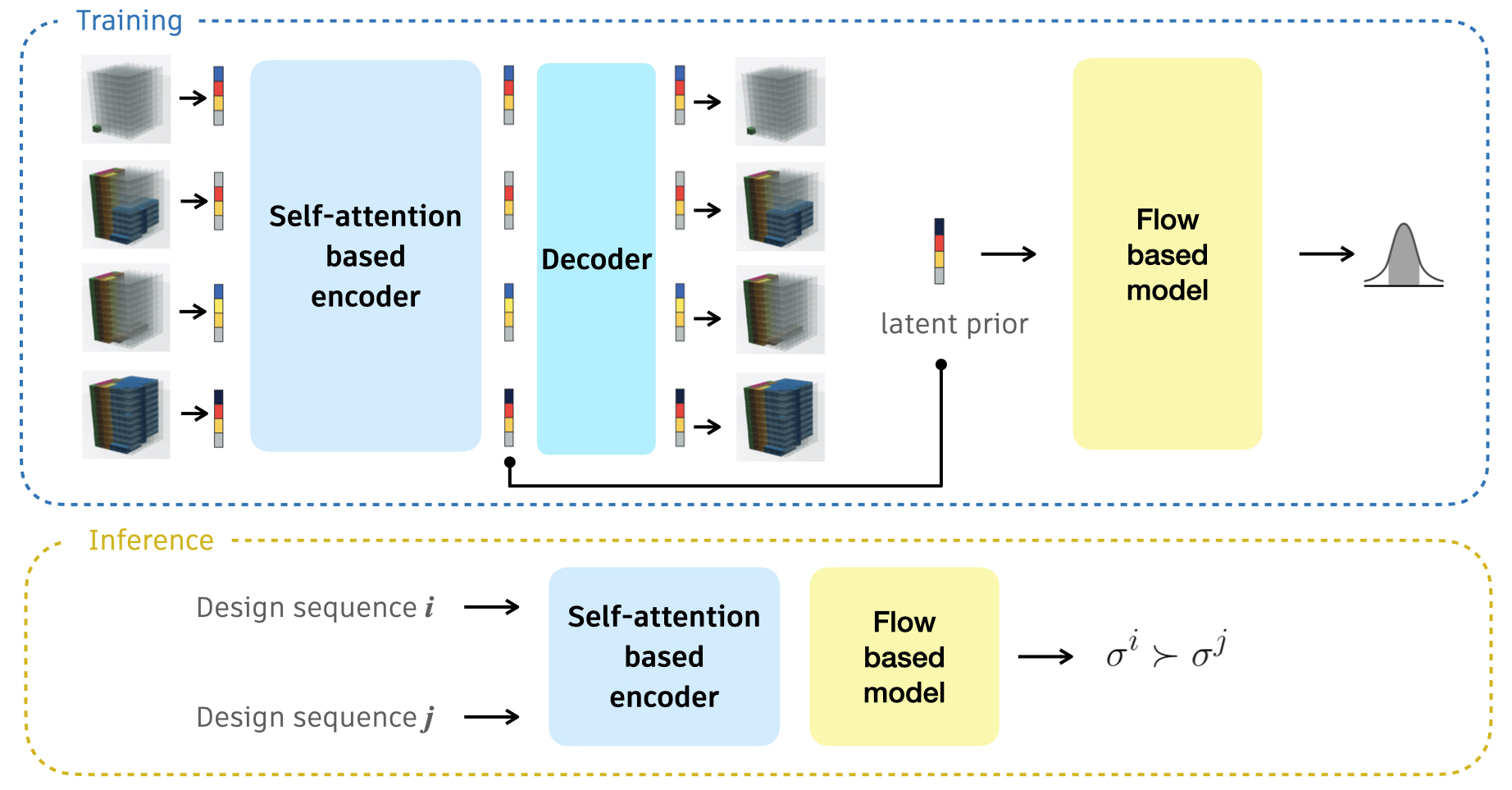}
    \caption{We develop an encoder-decoder based model for learning representations of sequential volumetric design data. The encoder portion consists of multi-head self-attention architectures. The output from the final self-attention layer is used to further train a flow-based model, e.g. real NVP for density estimation. Note that we keep the weights of the encoder frozen while training the flow model. During inference time, weights of both the encoder and the flow model are kept frozen to obtain the log-likelihood of a design sequence. Directly comparing this log-likelihood provides preference of design sequence $\sigma^i$ over design sequence $\sigma^j$.}
    \label{fig:my_label}
\end{figure*}

We train the transformer models using the reconstruction loss or the next design embedding prediction loss and extract the latent representations $(\mathbf{z}_0, \mathbf{z}_1, \dots, \mathbf{z}_T)$, where $\mathbf{z}_t\in\mathbb{R}^{2048}$, of each sequence $(\mathbf{e}_0, \mathbf{e}_1, \dots, \mathbf{e}_T)$ from the final self-attention layer. Positional encoding is also used before feeding the input to these attention layers. Next, we use the trained latent vectors as prior $p(\mathbf{z})$ to further train a flow-based model, e.g real NVP \cite{kingma2013auto}, for estimating the density. We choose flow-based models because they can work with high dimensional density estimation problems. Here we only use $\mathbf{z}_T$ to train the flow model although any number of latent vectors can be used by concatenating them. We consider that the latent code of the final state, $\mathbf{z}_T$, encodes information of the whole sequence because all of the previous states of a sequence contribute to this final and completed voxel design. Our proposed framework is illustrated in the training section of Fig. \ref{fig:my_label}. 

\subsection{Preference Model via Density Estimation}\label{sec:pref_model}
As an important downstream application of the learned representations, we develop preference models for volumetric designs. The core idea of a preference model is to assign preference between two input designs. This can be used in getting feedback to align a generative design model for desired user specifications. Our preference model consists of a pre-trained self-attention model and a pre-trained flow-based model. First, an input design sequence is passed through the pre-trained self-attention model. Second, output from the last attention layer is passed through the pre-trained flow-based model to achieve the log-likelihood. In this study, we only use the final latent vector of the sequence as it already contains information about the previous timesteps. In this way, we can obtain the log-likelihood of two input design sequences and compare them to provide preference. 
\begin{equation}
\text{preference}  = \begin{cases} 
                    \sigma^i \succ \sigma^j & \text{if  } \log p(\mathbf{z}_i) > \log p(\mathbf{z}_j) \\  
                    \sigma^i \prec \sigma^j & \text{if  } \log p(\mathbf{z}_i) < \log p(\mathbf{z}_j) 
                    \end{cases}
\normalsize
\end{equation}

\noindent Note that we use the notation $\sigma^i \succ \sigma^j$ to represent the preference of design sequence $i$ over design sequence $j$. Intuitively, this is equivalent to saying that the design sequence $i$ is more realistic than the design sequence $j$. The process is illustrated in the Inference section of Fig. \ref{fig:my_label}.  As a baseline, we also develop a simple VAE-based preference calculation procedure and compare it against the proposed preference model. For this baseline, we train a VAE by minimizing Eq. \ref{eq:elbo} which gives us the posterior of the latent space regularized towards the mean of the prior. For an input design sequence, we calculate the distance between the last latent vector of the sequence and the mean, $\mu$, of the prior. Next, we compare the distances between two input design sequences to identify the preference. 
\begin{equation}
\small
\text{VAE-preference}  = \begin{cases} 
                    \sigma^i \succ \sigma^j & \text{if  } ||\mathbf{z}_i - \mu || < ||\mathbf{z}_j - \mu || \\  
                    \sigma^i \prec \sigma^j & \text{if  } ||\mathbf{z}_i - \mu || > ||\mathbf{z}_j - \mu || 
                    \end{cases}
\normalsize
\end{equation}

\subsection{Building Design Autocompletion}
As a second downstream application of the learned representations, our goal is to feed a partial sequence to a model as input and obtain a completed sequence of a design. This happens in an auto-regressive manner where each design state depends on all the previous states of that design sequence and is thus predicted auto-regressively. Unlike language models, the raw output of the auto-regressive model cannot be used for volumetric designs as it is not possible to map the embedding to a specific design state. We identify this as a major challenge for generative architectural design models. To make sure that each generated design state follows the previous states, we use an additional mapping layer $\psi$, that ensures that no existing room from the previous state gets deleted although the room type can be changed. Note that when the model autocompletes the structure of the building, we allow it to change the room types for generating diverse designs that are different from the training data. If $Dec$ is the decoder, $Enc$ is the self-attention-based encoder, $\psi$ is the mapping layer, and $\mathbf{e}_t$ is the $t$-th design embedding of the sequence then we can write the following, $\mathbf{e}_t = \psi((\hat{\mathbf{e}}_t))$ where $\hat{\mathbf{e}}_t = Dec(Enc(\mathbf{e}_{t-1}))$. Finally, each embedding is converted back to a design state for voxel space representation. Evaluating these generated designs is challenging as we cannot compare them against the ground truth similar to the reconstruction task. To this end, we propose ``sequential FID'' scores for evaluating generated designs. This idea is an extension of the traditional image generation evaluation using the ``Fr\'echet Inception Distance''(FID) score \cite{heusel2017gans} in the sequential setting. The key idea is to obtain the Gaussian distribution of the $t$-th generated design embedding with mean, $\mathbf{m}_{w, t}$ and variance, $\mathbf{C}_{w, t}$ and compare them against the Gaussian distribution of the $t$-th design embedding from the training dataset with mean $\mathbf{m}_t$, and variance $\mathbf{C}_t$ using the following formula, 
\begin{equation}
\begin{split}
    f_t(& (\mathbf{m}_t, \mathbf{C}_t), (\mathbf{m}_{w, t}, \mathbf{C}_{w, t})) = ||\mathbf{m}_t - \mathbf{m}_{w, t}||_2 \\
    & + \text{Tr}\left(\mathbf{C}_t + \mathbf{C}_{w, t} - 2(\mathbf{C}_t \mathbf{C}_{w, t})^{1/2}\right).
\end{split}
\label{eq:seq_fid}
\end{equation}

\noindent We calculate the FID score sequentially for all of the design embeddings of the sequence.

\section{Experiment and Result}
\label{sec:experiments}

\subsection{Model Implementation and Training Details}
For the encoder part of our models, we use a GPT2-like structure that has several self-attention layers with multiple heads. Based on our experiments, we choose four self-attention layers and eight attention heads to run all of the experiments reported in this paper. We use a projection layer before the self-attention layers in our architecture. Additionally, the input sequence is masked so that future states are not available to the model. For the decoder part of our model, we use a simple linear layer and then a sigmoid layer to keep the predicted values between $0$ and $1$. To maintain the order of the sequence we use the sinusoidal positional encoding \citep{vaswani2017attention}. We train two different versions of this model: one for reconstructing each design embedding, which we call the `Volumetric Design Reconstruction (VDR)' model, and one for predicting the next design embedding from the current embeddings, which we call the `Autoregressive Volumetric Design (AVD)' model. Note that the VDR model predicts each current design embedding, $\hat{\mathbf{e}}_t$ from the corresponding input $\mathbf{e}_t$ and the AVD model predicts each next design embedding $\hat{\mathbf{e}}_{t+1}$ from the corresponding input $\mathbf{e}_t$. Each model is trained using stochastic gradient descent and Adam optimizer with a learning rate of $1\times 10^{-5}$ and binary cross-entropy loss. The latent dimension of the transformer-based model is $2048$. In the case of the flow-based model, we train a real-NVP flow model with $5$ coupling layers and a hidden dimension of $2048$ and use the Adma optimizer with a learning rate of $3\times 10^{-4}$. The batch size is $128$ in both cases.

\subsection{Reconstruction}
\begin{figure}[!htb]
    \centering
    \includegraphics[width=0.50\textwidth]{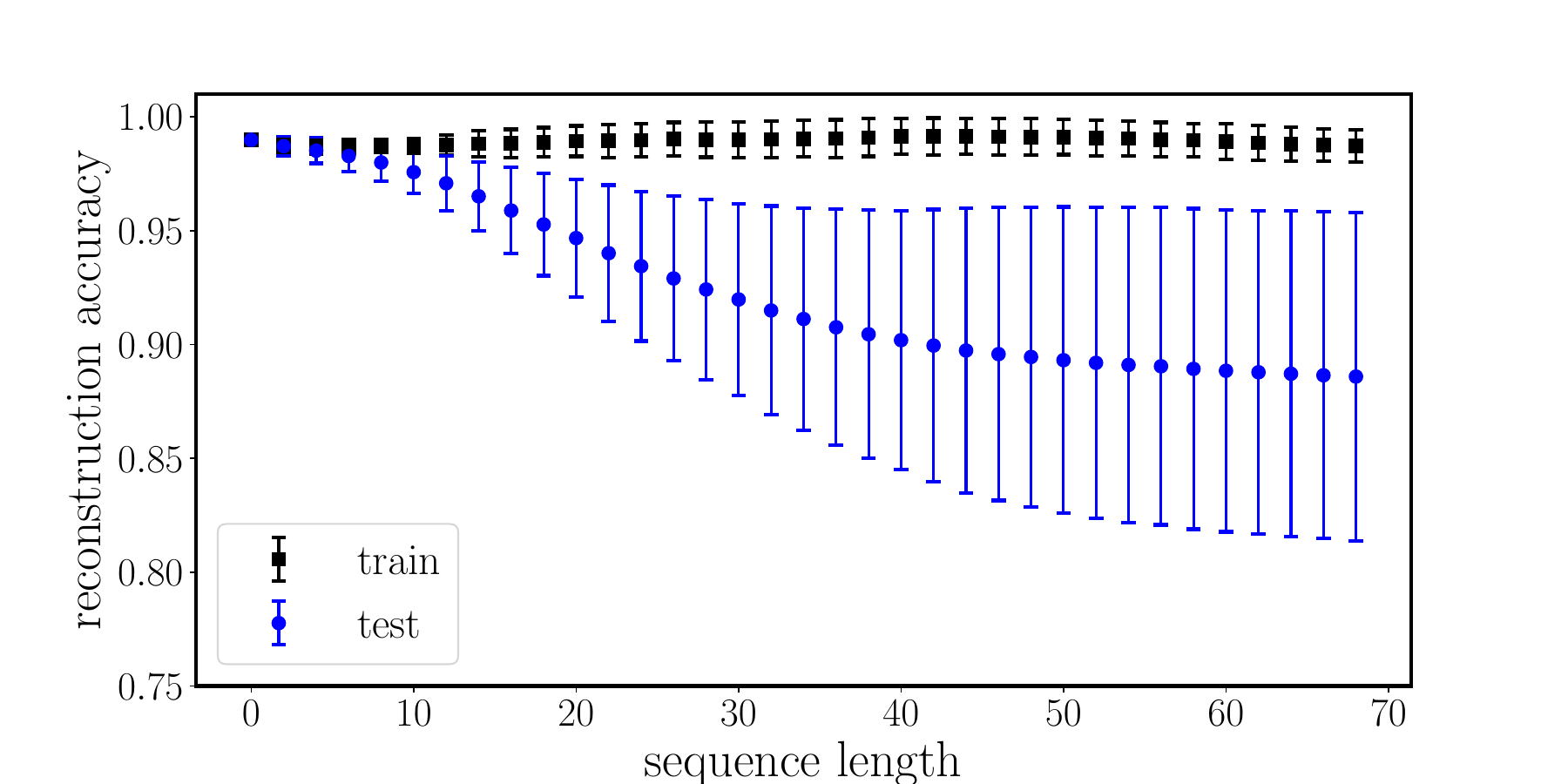}
    \caption{Reconstruction accuracy of training and evaluation dataset using the `Volumetric Design Reconstruction' (VDR) model}
    \label{fig:recons-accuracy}
\end{figure}

\begin{figure*}[!htb]
    \centering
    \includegraphics[width=0.80\textwidth]{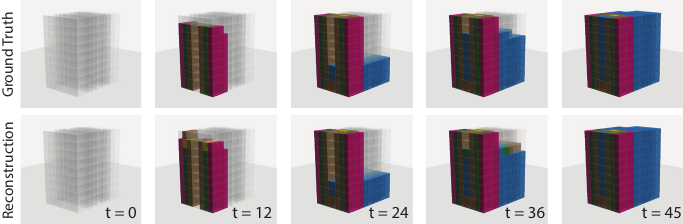}
    \caption{Reconstruction of a design sequence (below) from the training data and compare with ground truth (above).}
    \label{fig:recons_train}
\end{figure*}

\begin{figure*}[!htb]
    \centering
    \includegraphics[width=0.80\textwidth]{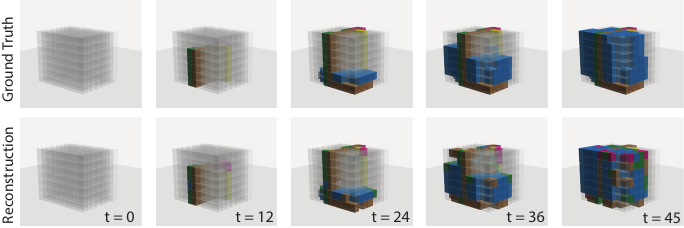}
    \caption{Reconstruction of a design sequence from evaluation data (below) and comparison with ground truth (above).}
    \label{fig:recons_test}
\end{figure*}

As described in the previous section, we train two models to learn the latent representations of the volumetric design data. The `Volumetric Design Reconstruction (VDR)' model is used to reconstruct each design embedding. Fig. \ref{fig:recons-accuracy} shows the reconstruction accuracy for both the training and evaluation datasets. As shown in the figure, the average reconstruction accuracy over the entire sequence is close to $100\%$ for the training data. This is impressive considering the fact that the model can generalize over a large range of volumetric design sequences. For the evaluation dataset, the model can achieve more than $80\%$ reconstruction accuracy, but it becomes less accurate over the longer timesteps. We hypothesize that including diverse designs in the training data may improve the evaluation accuracy and provide better generalization for design tasks. We conduct an ablation study on the number of attention layers and heads. The results are listed in Table \ref{tab:benchmark}. Different layers and heads don't show significant differences in the reconstruction accuracy. Finally, we visualize two sequences and their corresponding reconstructed designs using our VDR model in Fig. \ref{fig:recons_train} and \ref{fig:recons_test}. The qualitative results match the trend in Fig \ref{fig:recons-accuracy}, where some errors are introduced in the later steps from the evaluation data. However, the structure of the building and the majority of the room types are still preserved. More qualitative results can be found in the Appendix.  

\begin{table}
    \centering
    \small
    \begin{tabular}{c c c c}
    \hline
    \multirow{2}{*}{Layers} & \multirow{2}{*}{Heads} & \multicolumn{2}{c}{Reconstruction accuracy}\\
    \cline{3-4}
    & & Training & Evaluation\\
    \hline 
    $4$ & $2$ & $ 99.34\%+0.0063$ & $ 92.26\%+0.0656$ \\ 
    $4$ & $4$ & $ 99.24\%+0.0074$ & $ 92.11\%+0.0660$ \\
    $4$ & $8$ & $ 99.27\%+0.0069$ & $ 92.23\%+0.0661$ \\
    $8$ & $2$ & $ 99.34\%+0.0066$ & $ 89.71\%+0.0758$ \\
    $8$ & $4$ & $ 99.30\%+0.0071$ & $ 89.74\%+0.0758$ \\
    $8$ & $8$ & $ 99.26\%+0.0078$ & $ 89.80\%+0.0758$ \\
    \hline 
    \end{tabular} 
    \caption{Comparison between various model architectures for reconstruction accuracy for both training and evaluation dataset}
    \label{tab:benchmark}
\end{table}

\begin{figure*}[!htb]
    \centering
    \includegraphics[width=0.80\textwidth]{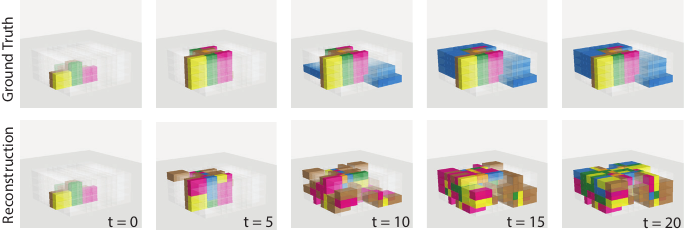}
    \caption{Autocompletion with a partial design given at t=0 from training data (below) and comparison with ground truth (above).}
    \label{fig:auto_train}
\end{figure*}

\begin{figure*}[!htb]
    \centering
    \includegraphics[width=0.80\textwidth]{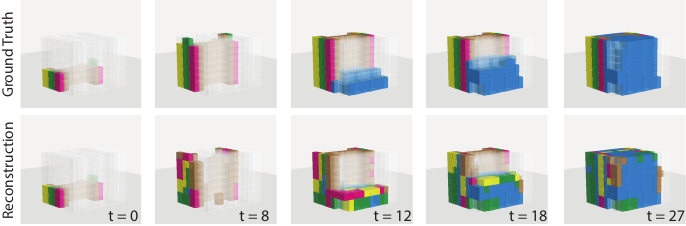}
    \caption{Autocompletion with a partial design given at t=0 from evaluation data (below) and compared with ground truth (above).}
    \label{fig:auto_test}
\end{figure*}

\subsection{Preference Model}
\begin{figure}
    \centering
    \includegraphics[width=0.475\textwidth]{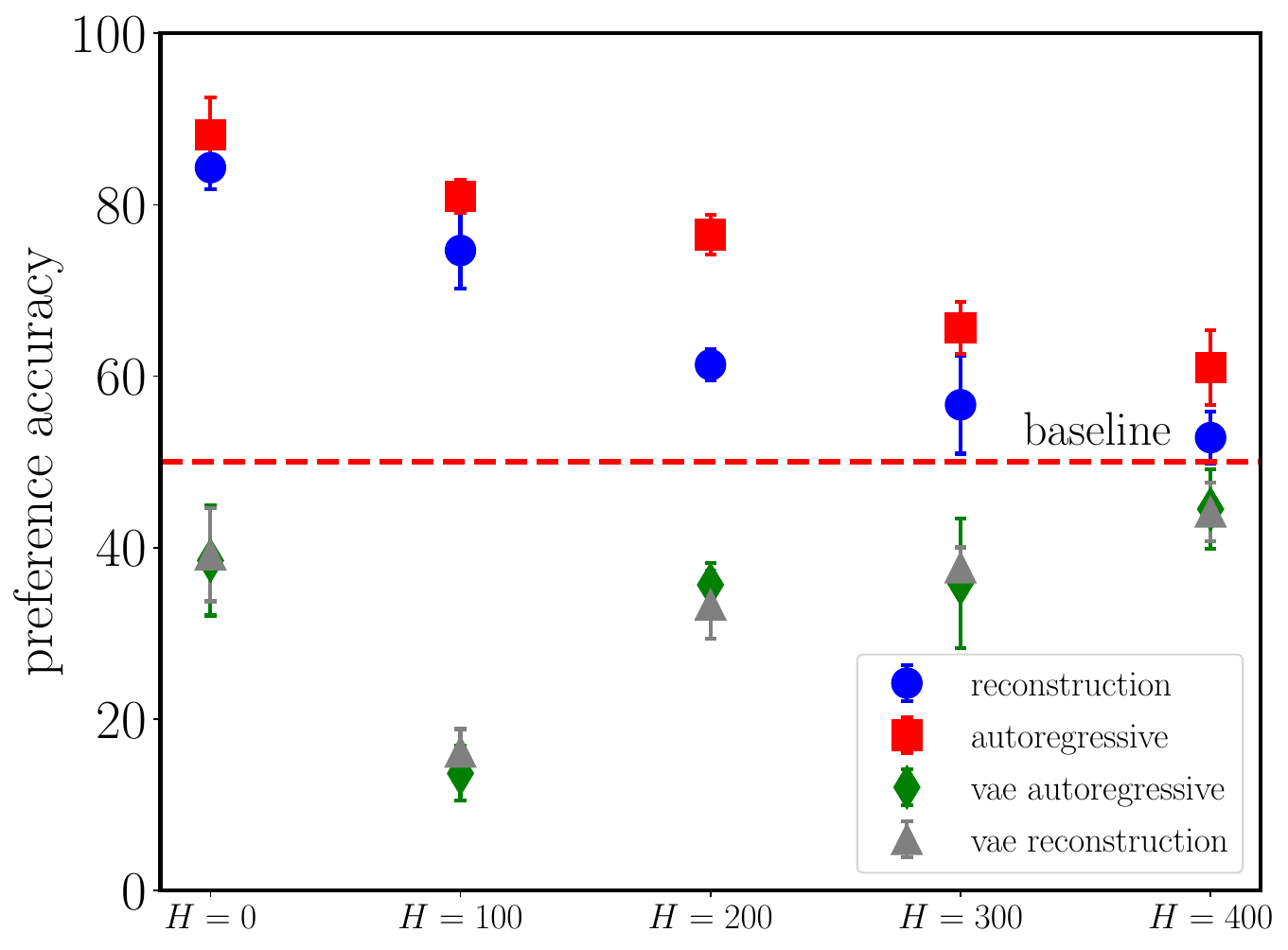}
    \caption{Preference accuracy of the Autogressive Volumetric Design (AVD) model and the Volumetric Design Reconstruction (VDR) model are shown with one standard deviation range, preference accuracy from the two types of VAE models are also shown; one for reconstruction and one for autoregressive prediction} 
    \label{fig:pref_score}
\end{figure}

To evaluate the performance of the preference model, we create five different versions of the evaluation datasets that each consist of $684$ sequences of volumetric designs. This is done by following the heuristic agent until a fixed horizon length, $H$, and then following a random policy for designs. This means that $H = 0$ are completely random designs, and $H = 400$ are close to expert designs. We choose $H = 0, 100, 200, 300, 400$ to gradually make the designs less random. Some examples of each evaluation dataset are shown in the appendix. The results of this test are shown in Fig. \ref{fig:pref_score}. As the designs become closer to the expert designs, the accuracy of our preference model decreases. This aligns with our expectations because the correlation within the design sequence becomes very close to the ground truth.

Additionally, we find that the learned representations from the AVD model provide better preference accuracy compared to the VDR model. This might be due to the fact that the autoregressive models can capture the correlation between design embeddings over the entire sequence, unlike the reconstruction model. The preference accuracy obtained using AVD is close to $90\%$ against completely random designs. For less random designs, the preference accuracy drops almost linearly. Note that for the $H=400$ dataset, the preference score is close to $60\%$. This can be explained by the fact that most design sequences become very close to the ground truth in this case and thus the preference becomes almost random. For the VDR model, we observe similar but slightly less preference accuracy. The highest preference accuracy is close to $85\%$ in this case and this happens against completely random designs. On the other hand, the preference accuracy of the VAE-based model is significantly lower than both the AVD and VDR models. This directly follows our intuitions from section \ref{sec:pref_model}. Intuitively, this represents the difficulties of learning a useful latent posterior using the ELBO loss in \ref{eq:elbo} for volumetric designs. 

\subsection{Autocompletion}
\begin{figure}
    \centering
    \includegraphics[width=0.50\textwidth]{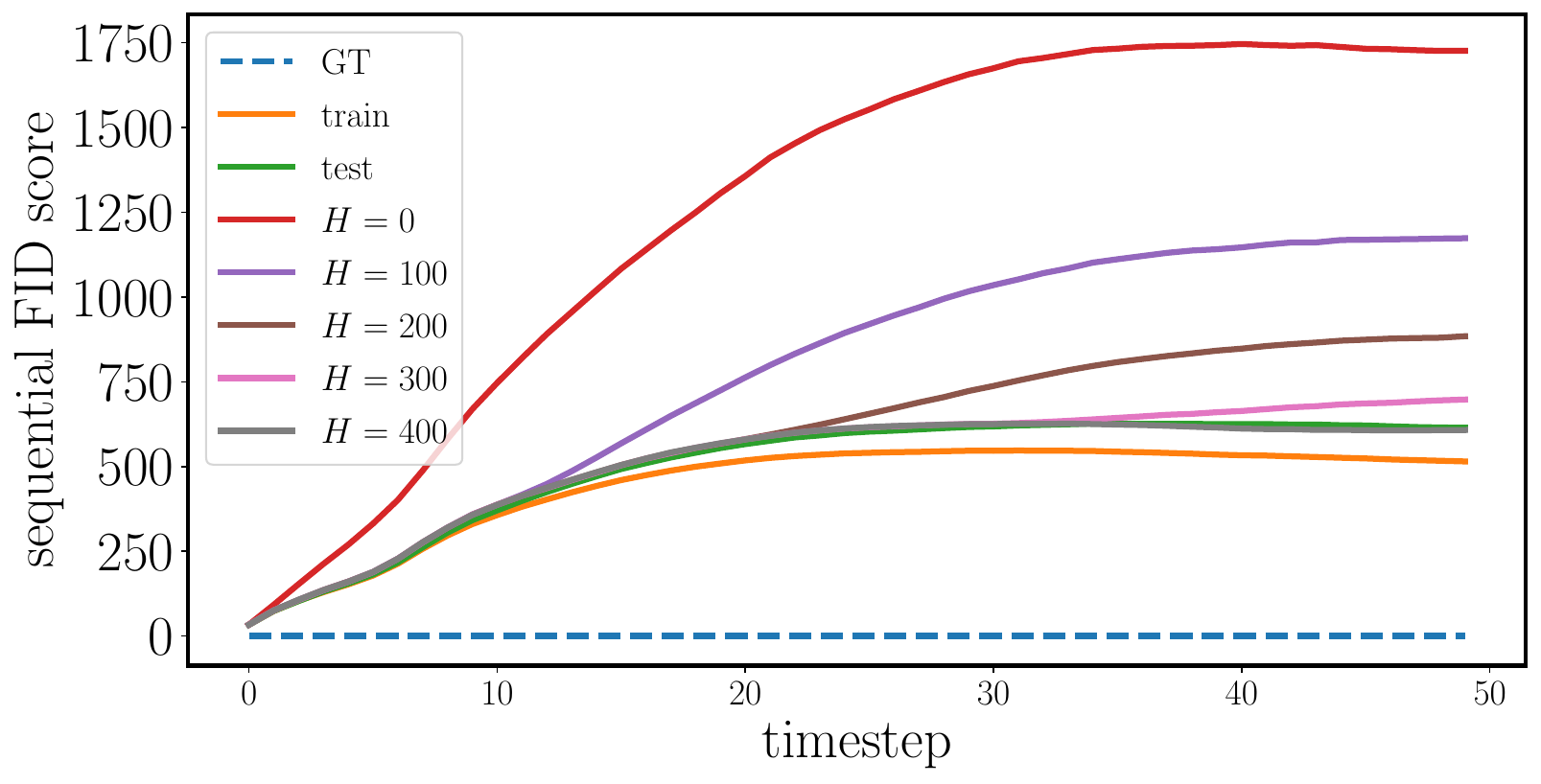}
    \caption{Comparing generated design sequences and random design sequences based on sequential FID scores to the ground truth for each timestep.}
    \label{fig:auto_perform}
\end{figure}

During inference time, we feed a partial input sequence of length $5$ to the AVD model and sequentially predict the next embeddings until a fixed timestep of length $50$. We demonstrate the qualitative design generation capability of AVD in Fig. \ref{fig:auto_train} and Fig.\ref{fig:auto_test} where the partial design input comes from the training and evaluation dataset respectively. For the first case, it is easy to notice that the autocompleted design follows the ground truth very closely with some minor variations. However, for the previously unseen input, the autocompleted design is diverse when compared to the ground truth. Although the structure of the design seems realistic, the room types are not assigned correctly for a realistic volumetric design. We suspect that this happens because the autoregressive model cannot properly capture the complex correlation between different room types. For a quantitative evaluation of the autocompleted sequences, we calculate the sequential FID score using equation \ref{eq:seq_fid}. We feed all the training data through the trained self-attention encoder and use the output latent vectors to calculate the mean and variance. This is shown in Fig. \ref{fig:auto_perform} where we also compare the sequential FID scores against various levels of randomness in the design sequences, $H=0, H=100, H=200, H=300, H=400$. The interpretation of this result is that the autoregressive model can generate designs that are more realistic than a random design sequence. The sequential FID scores are very close to the optimal design sequences, $H=400$, which justifies the credibility of the generated design sequences. Also, note that the FID scores increase for random design sequences which makes sense due to the randomness of each design state.

\section{Discussion \& Future Work}

We investigate the use of transformer-based models for learning representations of high-dimensional sequential volumetric design data. This is an important step toward developing data-driven models for evaluating and generating sequential design tasks. Our models can successfully reconstruct designs sequentially and accurately evaluate two design sequences with a trained flow-based density estimation model. We also examine auto-completion with our models, but the results are not satisfying. We believe this results from the complex spatial correlation between the room types in each volumetric design. As we sampled states from the original sequence to reduce the sequence length, it might have affected the sequential generation capabilities of the model. Additionally, the heuristic agent has limited data generation capabilities which also limits the generalization capacity of the trained model. Most importantly, the sequential design generation task is somewhat different from natural language processing tasks as there is no token associated with a building design state. Instead, our model is trying to directly predict the design embedding at each timestep. A potential solution might be to use carefully designed model architectures that can map the design embedding to a design state from a finite set of states. As this is one of the earliest studies on sequential design generation and design preference model building, we expect it to open new directions for research in this domain. Some future works include 1) developing metrics to better evaluate the quality of generated designs; 2) utilizing the preference model in a reinforcement learning framework to serve as a reward function engine, and 3) incorporating latent space searching methods (such as \cite{giacomello2019searching}) to not only generate plausible designs but also optimize for specific design objectives. The use of self-attention based architecture to capture sequential design decisions opens up possibilities of carrying over various methods and results from natural language processing to design domains. 

\section{Acknowledgement}

The authors would like to acknowledge the Autodesk AI Lab for their support. The authors would also like to acknowledge Linh Tran for providing useful guidance throughout the project and helping with the flow-based model.

\bibliographystyle{asmems4}
\bibliography{asme2e}

\newpage

\appendix       

\vspace{-0.5cm}
\section*{Appendix A: Details of Data Synthesis}
Together with professional architects, we design a heuristic agent to generate expert designs. Our heuristic agent is based on a particular representation of volumetric design called the voxel graph, a combination of voxel-based and graph-based representations proposed by \cite{chang2021building}. Note that a program graph is a type of graph that illustrates the relations between programs or rooms and is commonly used by professional architects for exploring design ideas. On the other hand, in a voxel graph, each node represents a voxel and the voxel information such as coordinate and dimension as node features. Most importantly, this representation allows non-uniform space partitioning and voids over-discretization using traditional uniform voxel size. The program type or room type is used as the node label. Based on randomly generated site conditions, we create the voxel graphs which are saved as json format. As the voxel graph is generated floor by floor, an action sequence can be computed that results in this voxel graph. Finally, we save the action sequence in json format. Note that Building-gym, our heuristic agent, takes this action sequence as input and generates the volumetric design sequentially. During each interaction with Building gym, the input is a particular action and output is a vector representation of the voxel state as described in \ref{sec:methodology}.1.

\section*{Appendix B: Heuristic Agent for Expert Design}

Here we describe the general principles of the heuristic agent:

\begin{enumerate}
  \item Generating an empty non-uniformed 3D partition space $[X, Y, H]$; 
  \item Based on the partition, FAR, TPR, determining the number of elevators;
  \item Positioning the elevators on the first floor to make sure they are symmetric to each other. For example, if only one elevator, then it should be placed in the center; if two elevators, they should be placed at two diagonal corners, etc. 
  \item Positioning service rooms (stairs, mechanical rooms, and restrooms) around the elevators; 
  \item Generating corridors to connect the above rooms;
  \item Duplicating the above layout to each floor to make sure they are vertically aligned; 
  \item Growing a lobby on the first floor until FAR \& TPR are satisfied; keeping the shape as rectangular as possible;
  \item Growing office rooms on the second floor until FAR \& TPR are satisfied; keeping the shape as rectangular as possible;
  \item Duplicating the layout of the second floor to the rest of the floors;
\end{enumerate}

\begin{table}[!htb]
\centering
\caption{Color coding for different room types used throughout this paper}
\begin{tabular}{c c}
     \hline 
     room type & color \\ \hline
     office & blue \\
     elevator & green \\ 
     stairs  & yellow \\ 
     restroom & pink \\ 
     Lobby & brown \\ 
     Mechanical & orange\\ \hline
\end{tabular}
\end{table}

\section*{Appendix C: Additional training details}
\begin{table}[h]
\begin{center}
\caption{Transformer model training details}
\begin{tabular}{c c}
     \hline 
     parameter & value \\ \hline
     learning rate & $1\times 10^{-5}$\\ 
     batch size & $128$\\ 
     epoch & $1000$\\ 
     number of self-attention layers & $4$\\ 
     number of attention heads & $8$\\ 
     attention dropout & $0.1$\\
     optimizer & Adam \\ 
     latent size & $2048$\\ \hline
    \end{tabular}
\end{center}
\end{table}

\begin{table}
\begin{center}
\caption{Flow model model training details}
\begin{tabular}{c c}
     \hline 
     parameter & value \\ \hline
     learning rate & $3\times 10^{-4}$\\ 
     batch size & $128$\\ 
     epoch & $1000$\\ 
     flow type & real-nvp\\ 
     number of coupling blocks & $5$\\ 
     optimizer & Adam \\
     scheduler & CosineAnnealingLR\\
     latent size & $2048$\\ \hline
\end{tabular}
\end{center}
\end{table}

\section*{Appendix D: Additional Dataset Statistics}
\renewcommand{\thefigure}{A\arabic{figure}}
\setcounter{figure}{0} 

\begin{figure}
    \centering
    \includegraphics[width=0.50\textwidth]{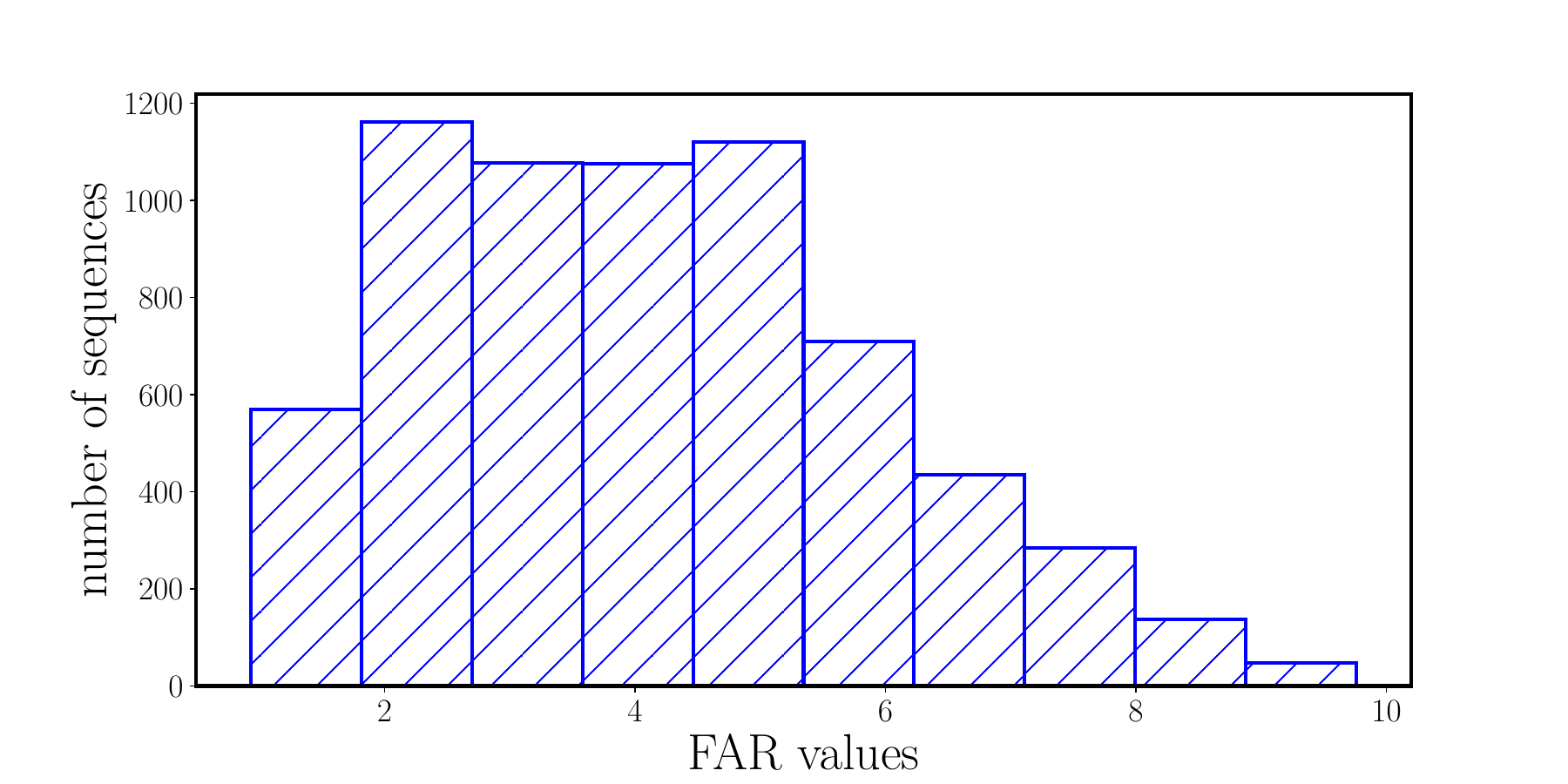}
    \caption{Distribution of FAR values in the volumetric design dataset}
    \label{fig:far_dist}
\end{figure}

\begin{figure}
    \centering
    \includegraphics[width=0.50\textwidth]{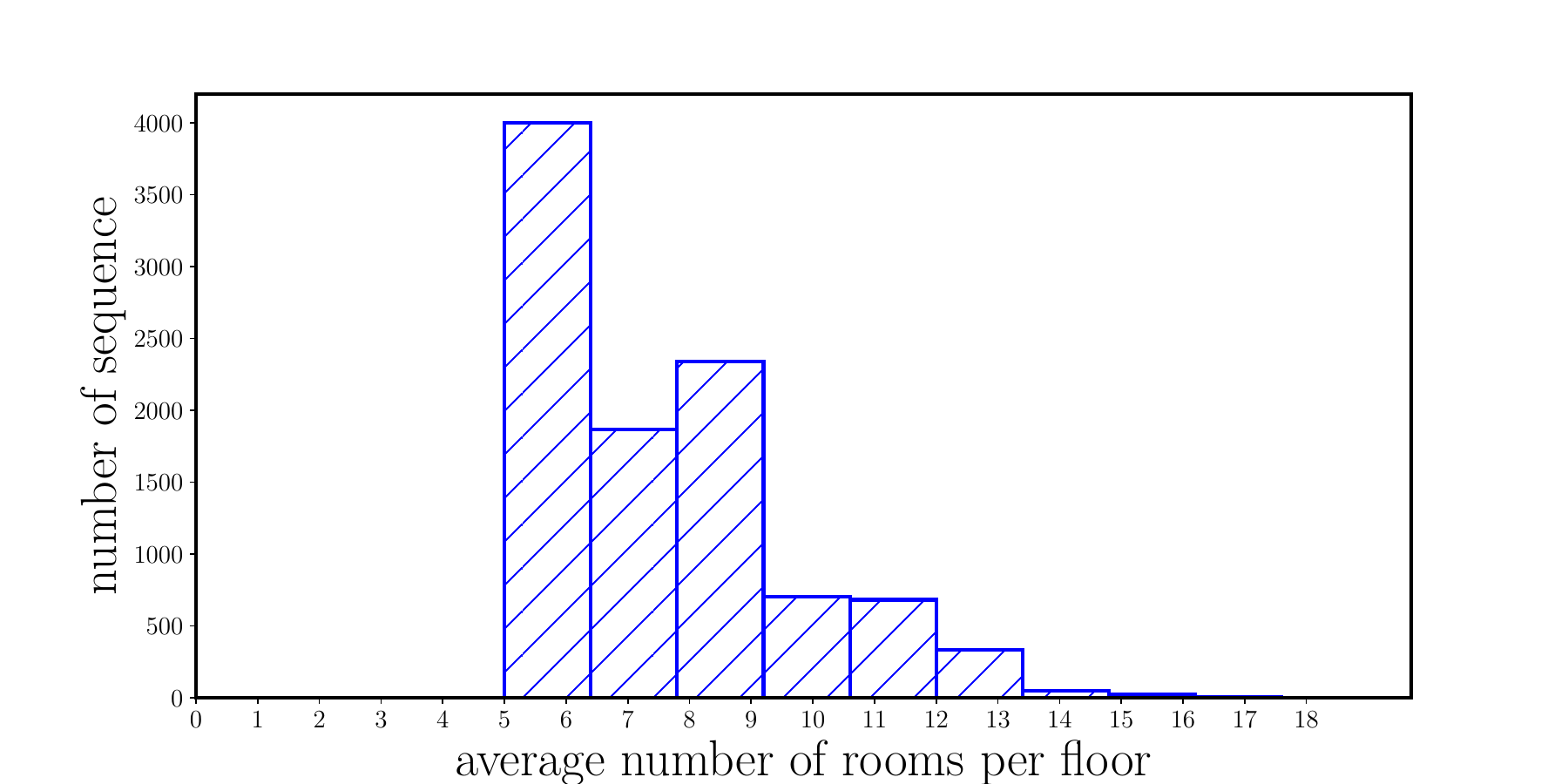}
    \caption{Distribution of the average number of rooms per floor in the volumetric design dataset}
    \label{fig:avg_room_num}
\end{figure}

Here we show the distribution of the floor-to-area (FAR) ratio values in Fig. \ref{fig:far_dist} and the distribution of the average number of rooms per floor in Fig. \ref{fig:avg_room_num}.

\section*{Appendix E: Additional Qualitative Results}
\begin{figure*}[!htb]
\centering
\includegraphics[width=1.0\textwidth]{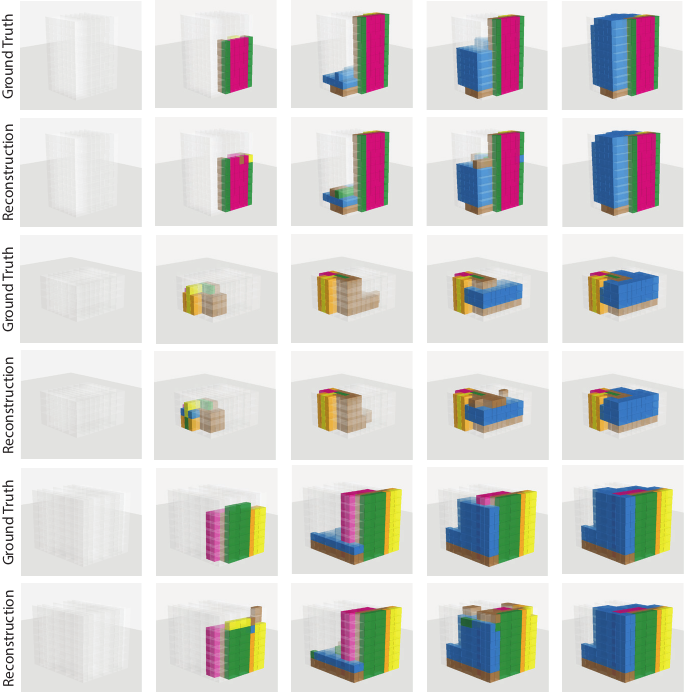}
\caption{\textbf{Reconstruction:} design reconstruction of sample sequences from the \textbf{training} data and comparison with the ground truth.}.
\label{fig:recons_train_2}
\end{figure*}

\begin{figure*}[!htb]
\centering
\includegraphics[width=1.0\textwidth]{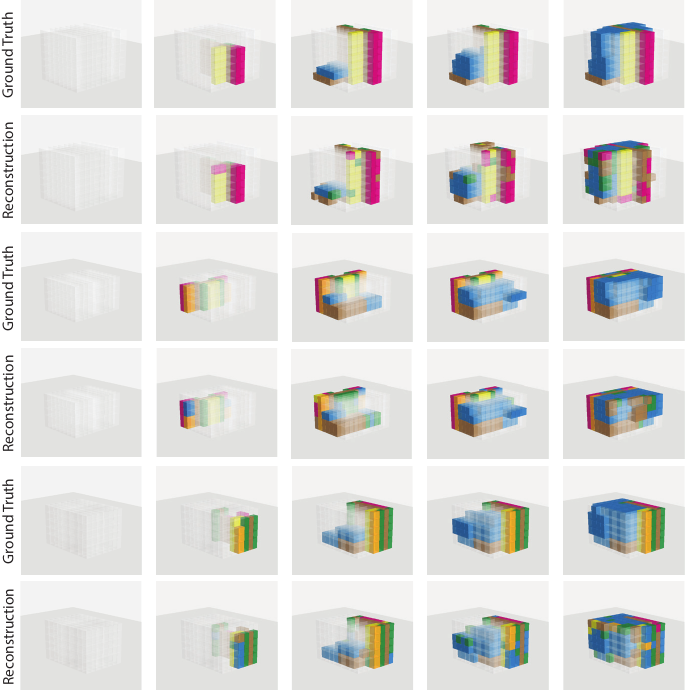}
\caption{\textbf{Reconstruction:} design reconstruction of sample sequences from the \textbf{evaluation} data and comparison with the ground truth.}.
\label{fig:recons_test_2}
\end{figure*}

\begin{figure*}[!htb]
\centering
\includegraphics[width=1.0\textwidth]{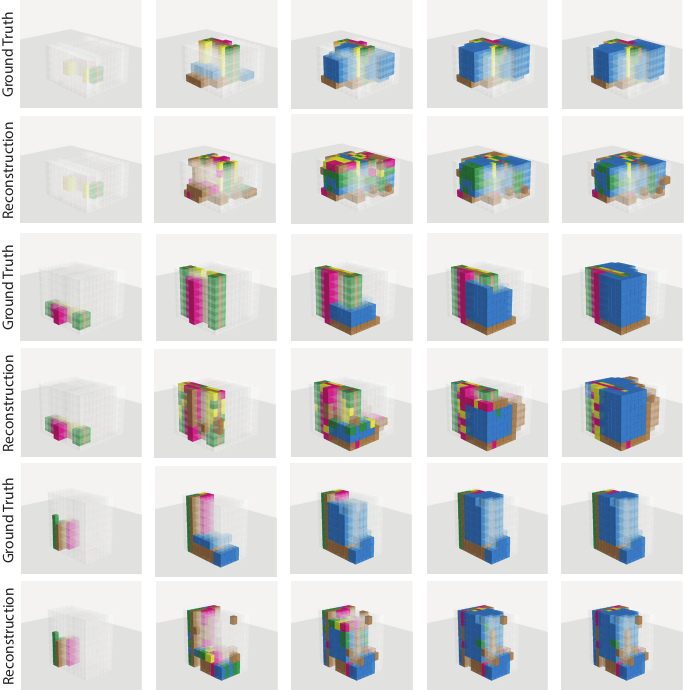}
\caption{\textbf{Autocompletion:} design autocompletion from a partial design given at t=0 from the \textbf{training} data and comparison with the ground truth.}.
\label{fig:auto_train_2}
\end{figure*}

\begin{figure*}[!htb]
\centering
\includegraphics[width=1.0\textwidth]{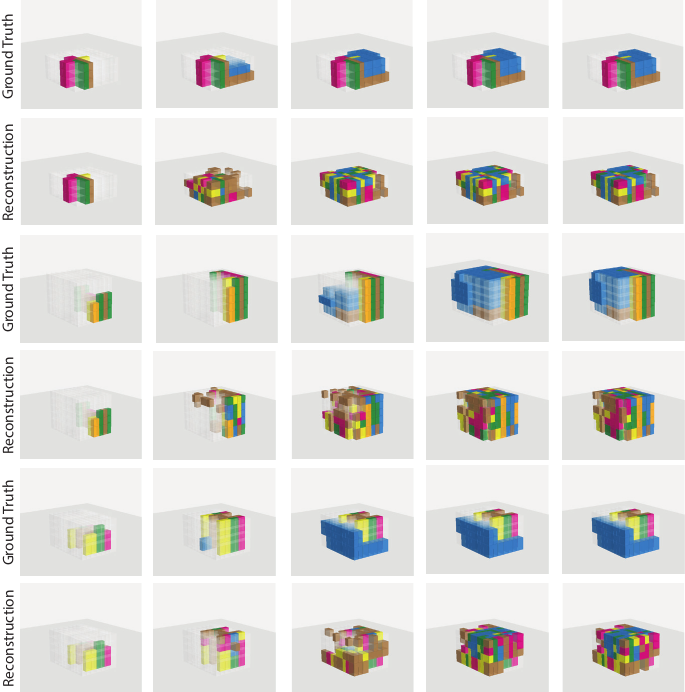}
\caption{\textbf{Autocompletion:} design autocompletion from a partial design given at t=0 from the \textbf{evaluation} data and comparison with the ground truth.}.
\label{fig:auto_test_2}
\end{figure*}

\begin{figure*}[!htb]
\centering
\includegraphics[width=1.0\textwidth]{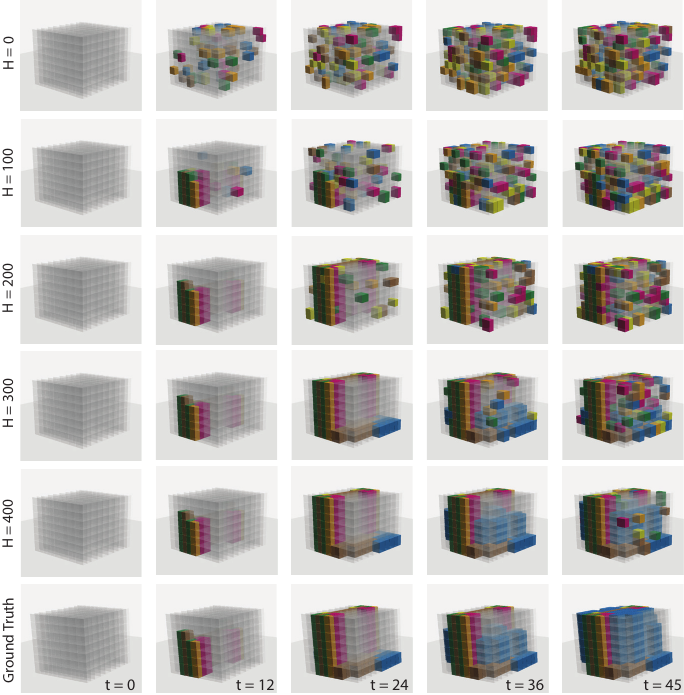}
\caption{An example sequence created from five different degrees of randomnesses of the evaluation data; a) $H=0$ means randomly generated sequence, b) $H=100$ means optimal policy being followed after 100 timesteps, c) b) $H=200$ means optimal policy being followed after 200 timesteps, d) $H=300$ means optimal policy being followed after 300 timesteps, e) $H=400$ means optimal policy being followed after 400 timesteps,}.
\label{fig:random}
\end{figure*}

Here we show some additional visualization of reconstructing sample sequences from both the training data and the evaluation data in figure \ref{fig:recons_train_2} and \ref{fig:recons_test_2}, respectively. We also show some autocompleted design sequences from partial input designs from both the training data and the evaluation data in figure \ref{fig:auto_train_2} and \ref{fig:auto_test_2}, respectively. We compare each design sequence against the ground truth. Finally, we visualize some sample sequences from the five different degrees of random datasets used to evaluate the preference model in figure \ref{fig:random}. 

\newpage 

\newpage

\newpage 

\newpage 

\end{document}